%% file: acl_latex.tex
\documentclass[11pt]{article}

\usepackage[preprint]{acl}

\usepackage{times}
\usepackage{latexsym}

\usepackage[T1]{fontenc}

\usepackage[utf8]{inputenc}

\usepackage{microtype}

\usepackage{inconsolata}

\usepackage{enumitem}

\usepackage{graphicx}
\usepackage{CJKutf8}
\input{math}

\usepackage{booktabs}
\usepackage{multirow}
\title{A Practical Evaluation Method for Long-Form Simultaneous Speech-to-Speech Translation}

\author{
 \textbf{Yulin Xue},
 \textbf{Siqi Ouyang},
 \textbf{Lei Li}
\\
 Carnegie Mellon University
\\
 \texttt{\{yulinx,siqiouya\}@andrew.cmu.edu,leili@cs.cmu.edu}
}

\begin{document}
\maketitle

\input{sections/abstract}
\input{sections/intro}
\input{sections/method}
\input{sections/experiment}
\input{sections/conclusion}

\bibliography{clean}

\end{document}

%% file: math.tex
\usepackage{amsmath,amsfonts,bm}

\def\eqref#1{equation~\ref{#1}}

\def\1{\bm{1}}

\def\vd{{\bm{d}}}

\def\vs{{\bm{s}}}
\def\vt{{\bm{t}}}

\def\vx{{\bm{x}}}
\def\vy{{\bm{y}}}

\DeclareMathAlphabet{\mathsfit}{\encodingdefault}{\sfdefault}{m}{sl}
\SetMathAlphabet{\mathsfit}{bold}{\encodingdefault}{\sfdefault}{bx}{n}



%% file: sections/abstract.tex
\begin{abstract}
Simultaneous speech-to-speech translation (SimulS2ST) enables real-time cross-lingual communication, but existing evaluation has focused largely on short or pre-segmented speech rather than long-form, continuous input. Prior approaches are difficult to reproduce and make assumptions that do not hold for end-to-end systems. We present a practical evaluation method for long-form SimulS2ST. Given source speech, pre-segmented source transcripts, and reference translations, we run automatic speech recognition (ASR) and forced alignment on the generated target speech to recover token-level timestamps, then apply a sentence-embedding-based aligner to match the target text to its corresponding source sentences. This enables sentence-level computation of latency and quality metrics, including YAAL and xCOMET, which are then aggregated into final system-level scores. Experiments on representative SimulS2ST systems show that the method is effective in practice and reveal that current systems suffer from substantial latency accumulation on long speech. Code can be found here \url{https://github.com/SakaiXue6666/Speech-to-Speech-Latency}.
\end{abstract}

%% file: sections/intro.tex
\section{Introduction}

Simultaneous Speech-to-Speech Translation (SimulS2ST) translates streaming source speech into target-language speech in real time~\citep{zheng-etal-2020-fluent}, enabling low-latency cross-lingual communication in scenarios such as multilingual conversations and international conferences. However, most prior work evaluates SimulS2ST on pre-segmented or short speech, despite the fact that real-world input, such as conference speech, is often continuous and may last for hours~\citep{sudoh2020simultaneousspeechtospeechtranslationneural,ma2022directsimultaneousspeechtospeechtranslation,liu22u_interspeech,communication2023seamlessmultilingualexpressivestreaming,zhang-etal-2024-streamspeech}.

One early effort toward long-form SimulS2ST evaluation is Boundary-Aware Latency (pBAL), which segments target speech into sentences, applies forced alignment to recover target token timestamps, and computes latency based on these timestamps~\citep{zheng-etal-2020-fluent}. This general paradigm is closely related to recent efforts in long-form simultaneous speech-to-text translation (SimulS2TT) evaluation~\citep{papi-etal-2024-streamatt,yaal}. However, pBAL has important practical limitations. First, it is not open-sourced, which makes it difficult to reproduce and adopt in subsequent research. Second, pBAL was designed for cascade systems comprising ASR, machine translation (MT), and text-to-speech (TTS), which introduces several limitations. In particular, it segments the target speech to align with the streaming ASR output of the source speech rather than with ground-truth source sentences, making the evaluation sensitive to source-side ASR errors. It also assumes access to target text for forced alignment, which is not available for some end-to-end (E2E) SimulS2ST systems~\citep{hibiki}.

In this paper, we propose a practical evaluation method for long-form SimulS2ST. We assume access to source speech, source transcripts pre-segmented into sentences, and their corresponding translation sentences. Given target speech produced by a SimulS2ST system, we first run ASR and forced alignment with state-of-the-art models to obtain target text with token-level timestamps. We then use the sentence-embedding-based method SEGALE~\citep{wang-etal-2025-extending} to segment the target text into sentences aligned with the source sentences. Finally, for each aligned sentence, we compute standard latency metrics such as YAAL~\citep{yaal} and quality metrics such as xCOMET~\citep{guerreiro-etal-2024-xcomet}, and average the sentence-level scores to obtain the final latency and quality scores. In our experiments, we evaluate several representative SimulS2ST systems with this method and analyze the quality of both ASR and sentence segmentation. We observe that even state-of-the-art systems exhibit latency accumulation on long speech. We will release the GitHub repository in the camera-ready version.

\section{Related Works}

\paragraph{Long-form simultaneous translation evaluation}
Latency evaluation for simultaneous translation has traditionally been studied in pre-segmented settings, where the input speech is split into utterances prior to evaluation.
StreamLAAL~\citep{papi-etal-2024-streamatt} extends utterance-level evaluation to the long-form setting by first segmenting the hypothesis into utterances aligned with the reference translation sentences using mwerSegmenter~\citep{matusov-etal-2005-evaluating}, then computing latency for each aligned hypothesis utterance and its corresponding reference sentence.
LongYAAL~\citep{yaal} improves upon StreamLAAL by mitigating the structural bias in latency evaluation and introduces SoftSegmenter, which yields better segmentation and alignment than mwerSegmenter.
These methods are designed for simultaneous speech-to-text translation, while our work extends them to the evaluation of simultaneous speech-to-speech translation.

\paragraph{Long-form machine translation evaluation.}
Another related line of work studies automatic evaluation for long-form machine translation.
mwerSegmenter~\citep{matusov-etal-2005-evaluating} aligns hypothesis and reference translation sentences by minimizing word error rate; however, it handles sentence boundaries poorly and often fails in cases of over- or under-translation.
SEGALE~\citep{wang-etal-2025-extending} improves upon mwerSegmenter by using a sentence boundary detector such as spaCy\footnote{\url{https://spacy.io/}} to recover sentence boundaries and by correctly penalizing over- and under-translation.
Our work leverages SEGALE as a more robust segmenter for the long-form hypothesis.

%% file: sections/method.tex
\section{Method}

In this section, we first introduce the formulation (Section~\ref{sec:formulation}). We then describe the ASR and forced alignment procedures (Section~\ref{sec:asr_fa}), the target speech segmentation method (Section~\ref{sec:segale}), and the computation of the final latency and quality scores (Section~\ref{sec:final}).

\subsection{Formulation}
\label{sec:formulation}

We define a long-form input speech stream as $\vs = (\vx_1, \vx_2, \cdots, \vx_n)$, where each $\vx_i \in \mathbb{R}^{|\vx_i|}$ denotes the speech waveform of the $i$-th sentence. Let $\vy_i$ denote the reference text translation of sentence $i$. Given the input speech stream $\vs$, a SimulS2ST system incrementally generates target speech $\hat{\vt}$. We assume that the input and target speech streams are temporally aligned at the start, i.e., they share the same initial timestamp. The goal of the evaluation method is to compute latency and quality scores for the generated target speech $\hat{\vt}$ given $\vs$, $\vx_{1:n}$, and $\vy_{1:n}$.

\subsection{Overview}
\label{sec:overview}

At a high level, our evaluation pipeline consists of three stages.
First, given the target speech generated by a SimulS2ST system, we run ASR to obtain the target-side text and apply forced alignment to recover token-level timestamps on the target speech.
Second, following SEGALE, we segment the target text into sentences and align them with the source transcript sentences and their reference translations, producing sentence groups that may reflect one-to-one, one-to-many, many-to-one, many-to-many, or null alignments.
Finally, for each aligned group, we compute latency using existing metrics such as YAAL and translation quality using sentence-level metrics such as xCOMET.
The group-level scores are then averaged into final system-level latency and quality scores.

\subsection{Transcribe with Timestamps}
\label{sec:asr_fa}

Given target speech $\hat{\vt}$, we use state-of-the-art ASR and forced alignment models: Qwen3-ASR-1.7B and Qwen3-ForcedAligner-0.6B~\citep{shi2026qwen3asrtechnicalreport}, to transcribe target speech $\hat{\vt}$ into text $\hat{\vy} = (\hat{y}_1,\cdots,\hat{y}_{|\hat{\vy}|})$ and obtain token-level timestamps $\vd = (d_1,\cdots,d_{|\hat{\vy}|})$ where $d_i$ denotes the end time of token $\hat{y}_i$. 

For long-form speech, we process the input in a chunk-wise manner. We divide the target speech $\hat{\vt}$ into $C$ consecutive chunks, each with duration 180 seconds,
\begin{align}
\hat{\vt} = (\hat{\vt}^{(1)}, \hat{\vt}^{(2)}, \ldots, \hat{\vt}^{(C)}).
\end{align}

For each chunk $\hat{\vt}^{(c)}$, the ASR model produces a partial transcription
\begin{align}
\hat{\vy}^{(c)} = (\hat{y}^{(c)}_1, \ldots, \hat{y}^{(c)}_{|\hat{\vy}^{(c)}|}).
\end{align}

Forced alignment is then applied to each chunk using the corresponding audio and recognized text to produce a chunk-level timestamp sequence
\begin{align}
\vd^{(c)} = (d^{(c)}_1, \ldots, d^{(c)}_{|\hat{\vy}^{(c)}|}),
\end{align}
where $d^{(c)}_i$ denotes the end time of token $\hat{y}^{(c)}_i$ within the $c$-th chunk. Let $o_c$ be the starting time offset of chunk $\vt^{(c)}$ in the original speech stream. We map chunk-level timestamps back to the global timeline by
\begin{align}
\tilde{d}^{(c)}_i = d^{(c)}_i + o_c.
\end{align}

Finally, the full transcription and timestamp sequence are obtained by concatenating all chunk-level results:
\begin{align}    
\hat{\vy} &= \hat{\vy}^{(1)} \oplus \cdots \oplus \hat{\vy}^{(C)} \\
\vd &= \tilde{\vd}^{(1)} \oplus \cdots \oplus \tilde{\vd}^{(C)}.
\end{align}

\subsection{Robust Segmentation with SEGALE}
\label{sec:segale}

We segment the target text $\hat{\vy}$ into sentence-level units and align them with the source speech sentences $\vx_{1:n}$ using SEGALE. We first split $\hat{\vy}$ into sentences $\hat{\vy}_{1:m}$ with spaCy~\footnote{\url{https://spacy.io/}}. Given the source speech sentences $\vx_{1:n}$, their reference translations $\vy_{1:n}$, and the segmented target sentences $\hat{\vy}_{1:m}$, SEGALE performs sentence alignment using Vecalign~\citep{thompson-koehn-2020-exploiting} with an adaptive skip-penalty search strategy.

To support many-to-many alignment, SEGALE constructs candidate contiguous spans on both the source and target sides, rather than restricting alignment to individual sentences. Let $\vx_{i:j} = \vx_i \oplus \cdots \oplus \vx_j$ denote a source span and $\hat{\vy}_{p:q} = \hat{\vy}_p \oplus \cdots \oplus \hat{\vy}_q$ denote a target span, where $\oplus$ denotes concatenation. For each source span $\vx_{i:j}$ and target span $\hat{\vy}_{p:q}$, SEGALE computes an embedding-based matching cost, with lower cost assigned to more semantically similar spans. Vecalign then finds a monotonic alignment between the source and target sentence sequences while allowing null alignments on either side, controlled by a skip penalty $\beta_{\mathrm{skip}}$.

The skip penalty determines the trade-off between forcing matches and allowing deletions. A large $\beta_{\mathrm{skip}}$ makes skipping expensive, so the aligner prefers fewer null alignments and more forced matches; this usually keeps the null-alignment ratio (NA ratio) low but increases the average alignment cost because semantically weak pairs are more likely to be matched. In contrast, a small $\beta_{\mathrm{skip}}$ makes skipping cheap, so the aligner more readily leaves segments unmatched; this typically increases the NA ratio and decreases the average alignment cost, since high-cost pairs are skipped and only easier matches remain. Therefore, SEGALE adaptively searches over $\beta_{\mathrm{skip}}$: it starts from a relatively large value and progressively decreases it in small steps. Once the average alignment cost falls below a threshold or the NA ratio exceeds a threshold, SEGALE treats this as the onset of over-deletion and returns the alignment from the previous step.

We denote the alignment output as
\begin{align}
A = (A_1, \ldots, A_r),
\end{align}
where each alignment group is defined as
\begin{align}
A_k = (X_k, Y_k, \hat{Y}_k).
\end{align}
Here, $X_k$ is a consecutive subset of source sentences from $\vx_{1:n}$, $Y_k$ is a consecutive subset of reference translation sentences from $\vy_{1:n}$, and $\hat{Y}_k$ is a consecutive subset of target sentences from $\hat{\vy}_{1:m}$.
SEGALE naturally handles both over-translation and under-translation. In the case of over-translation, some target sentences do not correspond to any source sentence, resulting in an empty $X_k$ and $Y_k$. In the case of under-translation, some source sentences do not correspond to any target sentence, resulting in an empty $\hat{Y}_k$. Such phenomena occur frequently in simultaneous translation, making this robustness important for long-form SimulS2ST evaluation.

\subsection{Latency Computation}
\label{sec:final}

Given the alignment produced by SEGALE, we compute latency at the alignment-group level. For each group $A_k = (X_k, Y_k, \hat{Y}_k)$, let $T_k^s$ and $T_k^e$ denote the start and end times of the source span $X_k$, respectively. Let $\vd_k = (d_1, \dots, d_{|\hat{Y}_k|})$ denote the token-level timestamps obtained by forced alignment for the target sentence group $\hat{Y}_k$.

We define the ideal delay of the $i$-th target token as
\begin{align}
d_i^{*} = T_k^s + (i - 1)\cdot \frac{T_k^e - T_k^s}{\max\{|Y_k|, |\hat{Y}_k|\}},
\end{align}
where $|Y_k|$ and $|\hat{Y}_k|$ are the numbers of tokens in the reference and target sentence groups, respectively.

The latency of group $A_k$ is then computed as
\begin{align}
l_k = \frac{1}{|\hat{Y}_k|} \sum_{i=1}^{|\hat{Y}_k|} (d_i - d_i^{*}).
\end{align}
Following LongYAAL~\citep{yaal}, we exclude the target tokens generated after the end of the full source stream $\vs$. 

Finally, we compute the long-form latency by averaging over all alignment groups:
\begin{align}
\mathrm{Latency} = \frac{1}{r}\sum_{k=1}^{r} l_k.
\end{align}
For over-translation or under-translation cases, where $X_k = Y_k = \emptyset$ or $\hat{Y}_k = \emptyset$, we exclude these groups from latency computation, since latency is not well-defined without both source and target content.

\subsection{Quality Computation}

Let $Q$ denote a sentence-level quality metric, such as COMET~\citep{rei-etal-2020-comet} or MetricX~\citep{juraska-etal-2024-metricx}. For each alignment group $A_k = (X_k, Y_k, \hat{Y}_k)$, we compute the quality score of group $A_k$ as
\begin{align}
q_k = Q(X_k, Y_k, \hat{Y}_k).
\end{align}

For over-translation or under-translation cases, we directly assign the minimum possible score of the metric, denoted by $Q_{\min}$. For example, $Q_{\min}=0$ for COMET and $Q_{\min}=-25$ for MetricX. The final long-form quality score is then computed by averaging over all alignment groups:
\begin{align}
\mathrm{Quality} = \frac{1}{r}\sum_{k=1}^{r} q_k.
\end{align}

%% file: sections/experiment.tex
\input{tables/acl6060}
\input{tables/antrex}
\input{tables/wer_cer}

\begin{figure*}
    \centering
    \includegraphics[width=1.0\linewidth]{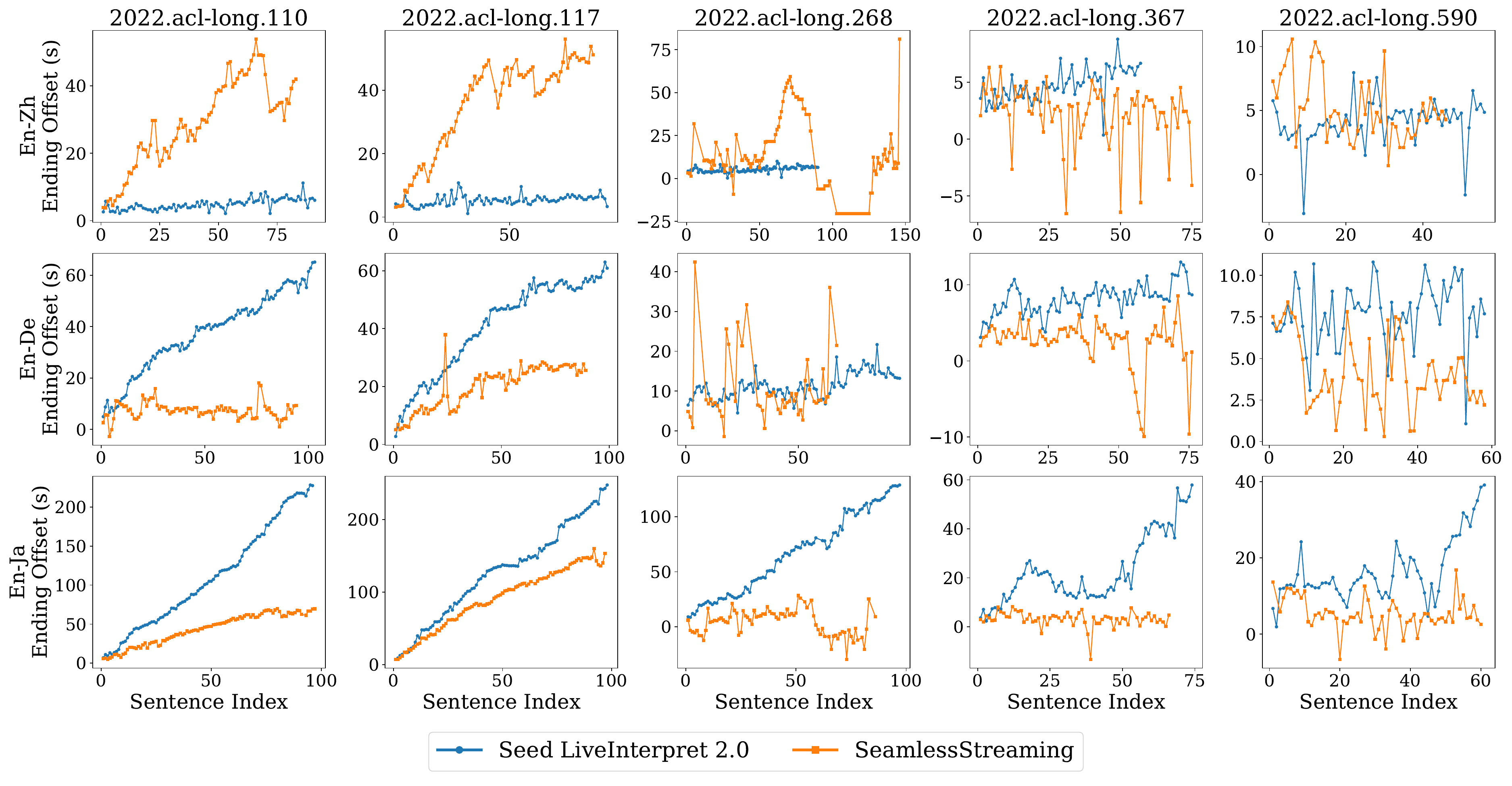}
    \caption{The ending offset of each aligned sentence for two systems on every speech in the ACL 60/60 dev set. The results show that latency generally accumulates as the source speech becomes longer.}
    \label{fig:latency_accumulation_acl}
\end{figure*}

\begin{figure*}[t]
    \centering
    \includegraphics[width=\textwidth]{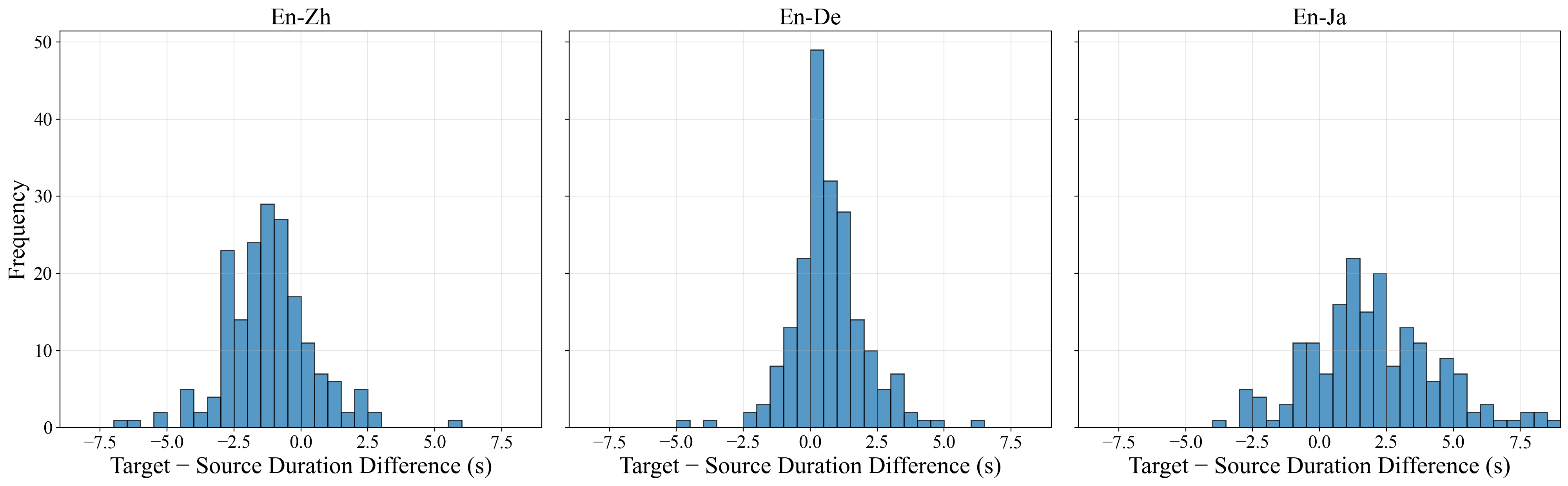}
    \caption{Distribution of target--source duration differences on the ACL 60/60 dev set. En$\rightarrow$Zh is mostly negative, En$\rightarrow$De is centered around zero, and En$\rightarrow$Ja is mostly positive.}
    \label{fig:tgt_minus_src_duration}
\end{figure*}

\section{Experiments}

\subsection{Setup}

\paragraph{Datasets} We evaluate existing SimulS2ST systems on two datasets: ACL 60/60 devv~\citep{salesky-etal-2023-evaluating} and Audio-NTREX-4L test~\citep{hibiki-zero}. ACL 60/60 dev consists of five English ACL talks, each approximately 10 minutes long, translated into multiple languages, in which we consider three directions: English to German/Japanese/Chinese. Audio-NTREX-4L is a multilingual speech translation benchmark introduced in Hibiki-Zero~\citep{hibiki-zero}. It is built from the NTREX text translation dataset by synthesizing source-language speech with high-quality TTS systems and multilingual speaker voices. The benchmark covers French, German, Portuguese, and Spanish as source languages and English as the target language. Each direction in our test split contains 450 speeches with an average duration of about 45 seconds, and we evaluate all four X-to-English directions.

\paragraph{SimulS2ST Systems} We consider three representative multilingual SimulS2ST systems:
\begin{itemize}[leftmargin=*]
    \item \textbf{Seed LiveInterpret 2.0}~\citep{cheng2025seedliveinterpret20endtoend}: a product-level end-to-end simultaneous interpretation system designed for high-fidelity, ultra-low-latency speech-to-speech generation. It supports voice cloning and is built on a duplex speech-to-speech architecture.
    
    \item \textbf{Hibiki-Zero}~\citep{hibiki-zero}: an end-to-end simultaneous speech-to-speech translation system built on the Moshi duplex architecture~\citep{moshi}. It is first trained on sentence-level aligned speech translation data and then further optimized with GRPO~\citep{shao2024deepseekmathpushinglimitsmathematical} to reduce latency while preserving translation quality.
    
    \item \textbf{SeamlessStreaming}~\citep{communication2023seamlessmultilingualexpressivestreaming}: a multilingual streaming speech translation model from the Seamless family. It uses Efficient Monotonic Multihead Attention (EMMA)~\citep{emma} to generate low-latency translations without waiting for the full source utterance, enabling simultaneous speech-to-speech and speech-to-text translation across multiple source and target languages.
\end{itemize}
Seed LiveInterpret 2.0 is evaluated through the Volcano Engine API~\footnote{\url{https://www.volcengine.com/docs/6561/1756902?lang=en}}, while Hibiki-Zero and SeamlessStreaming are run locally on a single NVIDIA L40S GPU.

\subsection{Evaluation}

The evaluation results on ACL 60/60 dev set are shown in Table~\ref{tab:acl6060_simuls2st}. 
Seed LiveInterpret 2.0 consistently achieves better translation quality than SeamlessStreaming, but at substantially higher latency, e.g., 9.4 seconds for En-Ja. 

The results on Audio-NTREX-L test set are shown in Table~\ref{tab:audio_ntrexl_simuls2st}. Overall, all systems achieve reasonably good translation quality. Among them, Seed LiveInterpret 2.0 obtains the best translation quality, but with more than 2 seconds higher latency than Hibiki-Zero and SeamlessStreaming.

We also observe that latency is less stable for En$\rightarrow$X directions, whereas it is much more consistent for X$\rightarrow$En directions.

\subsection{Analysis}\label{sec:analysis}

\paragraph{Poor En$\rightarrow$Ja Performance of Seed LiveInterpret 2.0} On En$\rightarrow$Ja direction, Seed LiveInterpret 2.0 exhibits very high latency, reaching nearly 10 seconds, while achieving an xCOMET-XL score of only 45.48. Our initial analysis suggests that the ASR transcripts of the generated target speech contain substantial gibberish and fragmented Japanese. To determine whether this issue stems from Qwen3-ASR's limited Japanese recognition performance or from poor Japanese speech synthesis quality, we measure word/character error rates (WER/CER) using both Qwen3-ASR-1.7B and WhisperX~\citep{bain23_interspeech}. The ground-truth target text is taken from the Seed LiveInterpret 2.0 API, which returns both synthesized target speech and target text. The results are shown in Table~\ref{tab:asr_acl6060}. Both Qwen3-ASR-1.7B and WhisperX yield very high CER on En$\rightarrow$Ja, suggesting that the problem is more likely caused by poor Japanese speech synthesis quality in Seed LiveInterpret 2.0.

\paragraph{Segmentation Quality} We analyze the segmentation quality of SEGALE on the ACL 60/60 En$\rightarrow$Zh dev set and compare it with the recently proposed SoftSegmenter~\citep{yaal}. The results show that SEGALE achieves a segmentation accuracy of 90.9\%, substantially outperforming SoftSegmenter (79.1\%). We observe that SoftSegmenter often shifts boundary-adjacent fragments across neighboring sentences, e.g., attaching the beginning of a sentence to the previous segment or the ending to the next one. Softsegmenter is also particularly brittle when semantically related content is realized with different surface forms. For example, when the reference contains foreign-language expressions while the prediction translates or paraphrases them, local token-level matching can break down. This is likely because its re-segmentation relies on local token-level matching rather than explicit sentence-level boundary modeling. In contrast, such errors are much less frequent with SEGALE. In contrast, SEGALE is much less affected by such cases and produces more semantically coherent segmentation.

\paragraph{Latency Accumulation} We further observe that, on long-form speech in the ACL 60/60 dev set (around 10 minutes), latency is substantially higher than on the Audio-NTREX-L test set (around 45 seconds). To better understand this phenomenon, we compute the ending offset of SEGALE-aligned sentences for both SeamlessStreaming and Seed LiveInterpret 2.0 on each speech in the ACL 60/60 dev set, as shown in Figure~\ref{fig:latency_accumulation_acl}. We find that on long speech, both systems exhibit increasingly larger ending offsets as more input speech arrives, with the only exception being Seed LiveInterpret 2.0 on the En$\rightarrow$Zh direction. 
Further analysis shows that latency accumulation is related to the sentence-level target-source duration difference, as shown in Figure~\ref{fig:tgt_minus_src_duration}. For En$\rightarrow$Zh, where the target speech is generally shorter than the source speech and the duration difference is concentrated in a negative range (roughly $[-2.5, -0.5]$ seconds), the ending offset stays small and stable. By contrast, for En$\rightarrow$Ja, where the target speech is typically longer than the source and the duration difference falls in a positive range (roughly $[0, 3]$ seconds), the ending offset exceeds 200 seconds near the end of the speech.
This suggests that even state-of-the-art SimulS2ST systems still suffer from latency accumulation on speech spanning minutes, highlighting the need for future research to address this issue.

%% file: tables/acl6060.tex
\begin{table*}[t]
\centering
\setlength{\tabcolsep}{5pt}
\begin{tabular}{lccc}
\toprule
System & En$\rightarrow$De & En$\rightarrow$Ja & En$\rightarrow$Zh \\
\midrule
SeamlessStreaming      & \textbf{4.333} / 67.56 & \textbf{2.434} / 42.89 & \textbf{1.725} / 40.66 \\
Seed LiveInterpret 2.0 & 7.939 / \textbf{85.39} & 9.413 / \textbf{45.48} & 5.306 / \textbf{72.78} \\
\bottomrule
\end{tabular}
\caption{Evaluation results of SimulS2ST systems on ACL 60/60 dev set. A / B denotes Latency (second) / xCOMET-XL. The best latency and quality scores are shown in bold.}
\label{tab:acl6060_simuls2st}
\end{table*}

%% file: tables/antrex.tex
\begin{table*}[t]
\centering
\setlength{\tabcolsep}{4.5pt}
\begin{tabular}{lcccc}
\toprule
System & Fr$\rightarrow$En & De$\rightarrow$En & Pt$\rightarrow$En & Es$\rightarrow$En \\
\midrule
SeamlessStreaming      & 3.520 / 77.50 & 3.833 / 78.95 & 3.566 / 76.11 & \textbf{3.608} / 77.88 \\
Seed LiveInterpret 2.0 & 5.892 / \textbf{86.67} & 5.933 / \textbf{88.63} & 5.530 / \textbf{86.94} & 5.592 / \textbf{88.63} \\
Hibiki-Zero            & \textbf{3.271} / 80.21 & \textbf{3.313} / 79.50 & \textbf{3.312} / 79.00 & 3.657 / 81.39 \\
\bottomrule
\end{tabular}
\caption{Evaluation results of SimulS2ST systems on Audio-NTREX-L test set. A / B denotes Latency (second) / xCOMET-XL. The best latency and quality scores are shown in bold.}
\label{tab:audio_ntrexl_simuls2st}
\end{table*}

%% file: tables/wer_cer.tex
\begin{table}[t]
\centering
\begin{tabular}{lccc}
\toprule
Model & De & Ja & Zh \\
\midrule
Qwen3-ASR-1.7B & 15.37 & 27.60 & 4.50 \\
WhisperX       & 13.80 & 27.30 & 5.52 \\
\bottomrule
\end{tabular}%
\caption{WER/CER of Qwen3-ASR-1.7B and WhisperX on generated target speech for ACL 60/60 dev.}
\label{tab:asr_acl6060}
\end{table}

%% file: sections/conclusion.tex
\section{Conclusion} 

We present a practical evaluation method for long-form SimulS2ST. Combining ASR, forced alignment, and SEGALE-based sentence alignment, it enables sentence-level evaluation of latency and translation quality on continuous speech across representative SimulS2ST systems. Experiments on ACL 60/60 and Audio-NTREX-L validate the method and show that SEGALE provides robust segmentation for long-form evaluation. More importantly, our analysis reveals a key limitation of current systems: latency accumulates substantially on long speech. We hope this work lays a strong foundation for future research on reliable, low-latency SimulS2ST.